\RequirePackage{fix-cm}
\documentclass[twocolumn]{svjour3}           

\smartqed  
\usepackage[pdftex]{graphicx}
\usepackage{graphicx}
\usepackage{cite}
\usepackage{amsmath,amssymb,amsfonts}
\usepackage{algorithmic}
\usepackage{textcomp}
\usepackage{wrapfig}
\usepackage{cite}
\usepackage{amsmath,amssymb,amsfonts}
\usepackage{adjustbox}
\usepackage{graphics}
\usepackage{textcomp}
\usepackage{booktabs}
\usepackage{multirow}
\usepackage{multicol}
\usepackage{amsmath}
\usepackage{xurl}
\usepackage{color, soul}
\usepackage{array}
\usepackage{tabularx}
\usepackage{changepage}
\usepackage{microtype}
\usepackage{subcaption}

\begin{document}

\title{Autobiasing Event Cameras for Flickering Mitigation
}


\author{Mehdi Sefidgar Dilmaghani\and Waseem Shariff \and Cian Ryan \and Joe Lemley \and Peter Corcoran}

\authorrunning{M.~Sefidgar Dilmaghani et al.} 

\institute{M. Sefidgar Dilmaghani \at
              University of Galway, University Road, Galway, Ireland\\
              \email{M.SefidgarDilmaghani1@universityofgalway.ie}           
           \and
          W. Shariff \at
              University of Galway, University Road, Galway, Ireland\\
              \email{W.Shariff1@universityofgalway.ie}           
           \and
          C. Ryan \at
              Tobii, Business Park, Parkmore East, Galway, Ireland\\
              \email{Cian.Ryan@tobii.com}           
           \and
          J. Lemley \at
              Tobii, Business Park, Parkmore East, Galway, Ireland\\
              \email{Joseph.Lemley@tobii.com}           
           \and
          P. Corcoran \at
              University of Galway, University Road, Galway, Ireland\\
              \email{Peter.Corcoran@universityofgalway.ie}           
}

\date{Received: date / Accepted: date}

\maketitle

\begin{abstract}
Understanding and mitigating flicker effects caused by rapid variations in light intensity is critical for enhancing the performance of event cameras in diverse environments. This paper introduces an innovative autonomous mechanism for tuning the biases of event cameras, effectively addressing flicker across a wide frequency range (25 Hz to 500 Hz). Unlike traditional methods that rely on additional hardware or software for flicker filtering, our approach leverages the event camera's inherent bias settings. Utilizing a simple Convolutional Neural Networks (CNNs), the system identifies instances of flicker in a spatial space and dynamically adjusts specific biases to minimize its impact. The efficacy of this autobiasing system was robustly tested using a face detector framework under both well-lit and low-light conditions, as well as across various frequencies. The results demonstrated significant improvements: enhanced YOLO confidence metrics for face detection, and an increased percentage of frames capturing detected faces. Moreover, the average gradient, which serves as an indicator of flicker presence through edge detection, decreased by 38.2\% in well-lit conditions and by 53.6\% in low-light conditions. These findings underscore the potential of our approach to significantly improve the functionality of event cameras in a range of adverse lighting scenarios.
\end{abstract}

\section{Introduction}
Event cameras have attracted significant attention in certain monitoring applications in recent years \cite{m20}. These sensors operate on an entirely distinct principle from frame-based cameras, taking inspiration from biological vision systems. As the ambient lighting conditions change, the camera detects change in the brightness of individual pixels asynchronously and generates a time-stamped stream of events. Compared to other types of cameras, this method offers several advantages, such as improved temporal resolution and a wider dynamic range. However, this improvement in temporal resolution can result in the generation of an excessive number of undesired events, especially in conditions characterized by flickering lighting \cite{m21}. 

"Flickering" in frame-based cameras and event-based sensors refers to the periodic variation in brightness or intensity caused by light source fluctuations, such as fluorescent or LED lights, interacting with the camera’s frame rate or exposure time. This interaction results in visual artifacts and inaccuracies in image processing tasks, such as object detection and tracking. In event-based systems, flickering triggers redundant sensor responses to non-existent changes in lighting or movement, not only impairing data analysis but also increasing processing time and power consumption. Furthermore, since all flickering events have the same amplitude, -1 or +1, distinguishing them from genuine events becomes challenging, complicating accurate data interpretation.

In recent years, research on the impact of flickering on camera output has been limited due to established methods for mitigating flickering in frame-based cameras and the reduced use of AC light sources in modern environments. However, flickering sources such as fluorescent and LED lights \cite{m11}, film and TV sets \cite{m12}, industrial machinery \cite{m13}, large screens, power lines, and modern vehicle systems still persist. In traditional RGB cameras, flickering from AC lighting was handled with a filter excluding 50/60 Hz frequencies. For event-based applications, the issue is more complex due to varying flickering frequencies and the lack of a fixed frame rate. Event cameras produce data as a raw event stream at different rates, making it impractical to use fixed frequency band filters as done previously \cite{m33}.

The integration of event cameras in everyday applications is a recent development, and challenges like flicker mitigation are still under-researched. \cite{m21} suggests a linear comb filter for flicker mitigation, but its fixed frequency and inability to operate in real-time limit its usefulness for real-time computer vision applications. To overcome the non-real-time limitation, \cite{m22} proposes using multiple filters for flicker elimination across 50 to 60 Hz frequencies, although this approach increases computational demands and requires specialized hardware. There is a need for a real-time method that can handle a broader range of frequencies and adapt to varying frame rates. Event camera manufacturers provide users the option to adjust specific sensor-level settings, known as biases, which control various aspects of the camera's output stream. This research aims to exploit these biases to dynamically adapt to different lighting conditions and mitigate flickering in real-time. It proposes a method to autonomously detect flickering and adjust the bias in the event camera, to remove flickering from the output event stream.

This research explores the development of an autobiasing system for event cameras in computer vision applications. A convolutional neural network (CNN) detects the presence of flickering, and if detected, a feedback loop dynamically adjusts the event camera’s internal settings , named bias, to mitigate it from the output event stream. This approach is tested on a YOLO-based face detection algorithm, and accurately detected faces even in low light conditions (under 10 lux) and in the presence of various flickering frequencies.

In this paper, we propose a novel algorithm that, for the first time:

\begin{itemize}
    \item Utilizes CNNs to detect the presence of flickering in the output of event cameras,
    \item Employs an automatic and dynamic feedback loop to tune the event camera's built-in bias settings to mitigate flickering,
    \item Assesses the efficacy of real-time autobiasing in flicker mitigation using the average gradient as a quantitative metric.
\end{itemize}

The structure of this paper is organized as follows: Initially, there will be a brief introduction to event cameras and bias settings. The next section will present a literature review that explores flickering mitigation techniques for both frame-based and event cameras, as well as an examination of event camera bias settings. This is followed by the methodology section, which will outline the research problem and proposed methods. The results and discussion section will then detail the outcomes of the tests and their interpretation. The paper concludes with a section on future work and final conclusions.

\section{Event Cameras Hardware and Biases}
\begin{figure}[b]
	\centering
		\includegraphics[width=0.8\linewidth]{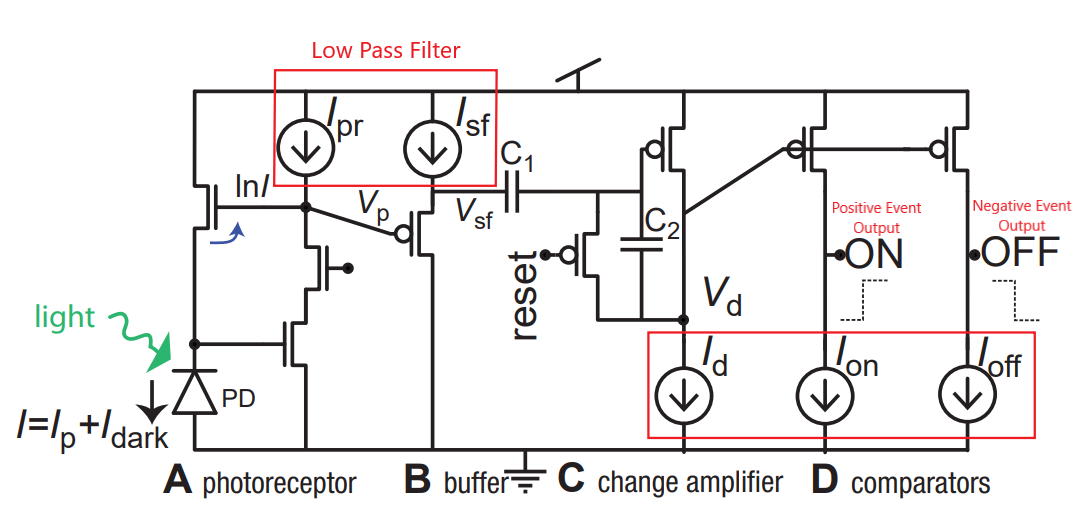}
	\caption{The hardware of each pixel in an event camera \cite{m10}.}
	\label{fig_6}
\end{figure}

This section looks into the hardware components of the event camera in more detail. The schematic of an event camera is shown in Figure \ref{fig_6}. The photodiode (PD), detects variations in the light in the environment and transforms these variations into an electrical current. After passing through a number of steps, this current finally triggers an event in the output. The buffer, shown as component 'B' in the schematic, is the part that makes this progress possible. Each event is followed by a reset of the system, which involves charges and discharges through C1 and C2. The I$_{\text{on}}$ and I$_{\text{off}}$ current sources calibrate the camera's sensitivity to either positive or negative light changes to complete the procedure. The aforementioned blocks are not locked to certain values and may be adjusted by the user via bias controls provided by the camera makers. Since each application has particular requirements, biases should be fine-tuned to get the optimum results. The low-pass filter shown in Figure \ref{fig_6} controls the buffer bandwidth, and event camera manufacturers provide a specific bias to adjust it. This bias helps reduce flickering effects in the event output, and this study specifically focuses on tuning it.

\subsubsection{bias\_fo (buffer bandwidth)}
The strategy taken in this study is to address the challenge of flickering as a single-variable issue, focusing on the variable with the highest impact on flickering mitigation. As stated, this particular bias directly controls the low-pass filter, thereby significantly influencing the reduction of flicker effects more than other biases. In the camera used in this project, this bias is referred to as bias\_fo. However, other event cameras also have similar capabilities, so this method is not specific to the camera used here. Other biases will remain at their default settings, as the exploration of adjustments to these is beyond the scope of this research. Attempting to tune all biases together would introduce a complex multi-input, multi-output optimization task, requiring advanced mathematical optimization techniques that diverge from the streamlined focus of the current project.

The buffer component depicted in Figure \ref{fig_6} functions analogously to a second-order low-pass filter. The cut-off frequency of the first stage is controlled by $I_{\text{p}}$ and $I_{\text{pr}}$, but the cut-off frequency of the second stage is controlled by bias\_fo, which is attained by adjusting $I_{\text{sf}}$ level. 

Narrowing the range of this value can potentially enhance the suppression of low-frequency noise. However, it is crucial to recognize that such a modification may also lead to decreased sensor operational speed \cite{m39} and potential loss of real data. Thus, there is a trade-off between sensor speed and low-frequency noise mitigation that must be carefully managed. Therefore, it is recommended to initially set the bias\_fo towards the higher end of its range, reducing it only when necessary. Reduction should cease as soon as the flickering no longer disrupts the event based application which is face detection here. This approach ensures that an adequate amount of real data is maintained while effectively eliminating flickering.

\section{Literature Review}
\subsection{Flicker Mitigation in Frame-based Cameras}
The challenge of mitigating flickering artifacts in monitoring systems has been a persistent concern, especially in applications like driver monitoring. Conventional camera sensors, which have long served these domains, are susceptible to flickering issues due to variations in ambient lighting conditions or artificial light sources. Some of the hardware based flickering mitigation techniques are aligning the camera's frame rate with the local power frequency, utilize a global shutter to eliminate rolling shutter effects, and synchronize lighting equipment with the camera's frame rate. Alternative approaches employ continuous lighting sources, external lighting controllers, and power conditioners to stabilize lighting conditions \cite{m140}, \cite{m141}.  

Some of the digital image processing based flickering mitigation techniques are next outlined. One common approach to dealing with flickering in conventional cameras is frame averaging. This technique involves capturing multiple frames and averaging them to reduce flickering effects caused by rapid fluctuations in lighting conditions. Kim et al. \cite{m14} proposed a method that employs frame averaging to enhance image quality in low-light surveillance scenarios. However, this method is limited in its ability to handle rapidly changing flickering patterns. Several more notable deep neural network based contributions stand out. First, the work by G. Lu et al. \cite{m15} introduces a deep non-local Kalman network that effectively reduces compression artifacts through a combination of model-based and learning-based approaches. Lei C et al.\cite{m16} introduces a practical "blind deflickering" technique, valuable for its ability to remove flickering without specific guidance. Y. Xu et al.\cite{m17} approach to video inverse tone mapping offers a unique perspective on flicker reduction by considering the LDR-to-HDR transformation, showcasing the power of 3D convolutional networks. Paul S et al. \cite{m18} highlight the challenges of video analytics accuracy fluctuations and suggest transfer learning as a solution, indirectly addressing flickering concerns. Lastly, Pony R et al. \cite{m19} discuss adversarial flickering attacks, underscoring the importance of robust flicker mitigation techniques in video recognition systems. 

As stated, recent literature provides an in-depth exploration of conventional camera techniques for mitigating flickering, incorporating a range of methods from deep learning to transfer learning. However, it's important to note that while those approaches offer potential solutions for conventional cameras, they may not translate effectively to event camera-based systems. Common issues in these works include their design for fixed frequencies and the need for separate filters to remove flickering.

\subsection{Flickering in Event Camera}
Recent research aimed at addressing the effects of lighting flicker on event cameras has seen a variety of innovative approaches. Wang et al. \cite{m21} address flicker sensitivity by introducing a linear comb filter that preprocesses event data, significantly improving the signal-to-noise ratio. This is particularly beneficial for robotics applications in indoor settings with flickering light sources. However, this work is not designed for real-time event data processing and is limited to a fixed frequency. Im et al. \cite{m22} present PINK, a real-time and lightweight algorithm that filters flicker based on event polarity, demonstrating efficacy in extreme flicker environments and providing usable optical flow results for machine vision applications. Again, this work is tested only on 50 Hz and 60 Hz frequencies. Muglikar et al. \cite{m25} propose a neural network-based calibration framework for event cameras, which eliminates the need for active illumination and enables intrinsic and extrinsic calibration under common distortion models.

Tang et al. \cite{m26} introduce a denoising method based on salient region recognition, effectively removing background and irregular noise from event camera outputs caused by flickering light sources. Although high-frequency flickering is mentioned as a problem in their work, their tests are conducted on additive noise and there is no information provided about the proposed system's performance under various flickering frequencies. Xu et al. \cite{m27} explore event-based Electric Network Frequency (ENF) estimation, surpassing traditional video-based methods in challenging environments through mode filtering and harmonic enhancement. \cite{m24} is not focused on flickering mitigation; rather, it proposes a technique to combine various flickering frequencies under bright light conditions in an event camera to develop a projector-event camera system. This proposed method offers three main advantages over previously mentioned methods: it is not frequency-fixed and performs well across a range of frequencies from 25 Hz to 500 Hz, it operates in real-time, and it eliminates the need for additional hardware design and implementation. It should also be noted that there are currently no datasets available to evaluate the impact of such flickering using biases.

\subsection{Event Camera Tuning and Biases}
The field of event cameras has seen significant advancements in bias configurations, shaping the performance and adaptability of these sensors. Lichtsteiner et al. \cite{m28} introduced a CMOS vision sensor with high pixel bandwidth and wide dynamic range, providing low-latency dynamic vision under varying illumination. Graca and Delbruck \cite{m29} unraveled the paradox of intensity-dependent dynamic vision sensor (DVS) pixel noise, revealing the DVS photoreceptor as a second-order system. Graca et al. \cite{m30} explored optimal biasing, establishing theoretical limits and showcasing the effectiveness of high photoreceptor bias. Delbruck et al. \cite{m10} introduced fixed-step feedback controllers for automatic bias control, ensuring regulated event rates and managed noise. McReynolds et al. \cite{m31} proposed experimental methods to predict DVS performance, bridging the gap between sensor input and real-world output. Dilmaghani et al. \cite{m1} contributed by examining bias effects on output sharpness, providing insights into the relationship between bias configurations and event data quality. Overall, these studies collectively enhance our understanding of event camera biases, paving the way for improved performance and adaptability in various event applications.

\section{Methodology}
Event cameras record changes in light in real time instead of using typical frame-based approaches; therefore, they offer a promising solution for applications that need high-detail collection of information, like face tracking and blink counting. However, there are still difficulties, particularly when adjusting to different lighting situations. Noise sources, flickering or flashing lights, and low light levels can all have a negative effect on the output quality. The severity of the situation is increased by the fact that noise and data are displayed in event cameras with the same amplitude. These issues become more complex when there are several undesirable lighting conditions present at once. One particularly harmful scenario is when a flickering source operates in a dimly lit area. That affects vital applications like driver monitoring systems, where a face detection failure can have catastrophic consequences, particularly when the system is operating at night.

This issue can be resolved by developing a real-time method to mitigate flickering in event cameras without placing an excessive computational load on their processing units. Previous methods are considered insufficient since event cameras and traditional frame based cameras differ fundamentally, which calls for a more in-depth comprehension of event camera dynamics. Moreover, rather than employing fixed-frequency filters, proposed solutions should be adaptable enough to handle a wide range of flickering frequencies. Some control over the output stream can be achieved by utilizing bias settings; as tuning bias\_fo can lead to minimization of low-frequency background noise, and flickering.

To address the flickering issue, this study proposes an algorithm that keeps track of the event camera's output as well as the functionality of event-based applications. By automatically modifying biases, the algorithm maximizes performance. The main goal is to develop and implement an adaptive algorithm that can respond to changing environmental conditions in a dynamic manner, thus improving the efficiency and reliability of the event-based systems.

\begin{figure}[h]
	\centering
		\includegraphics[width=.85\linewidth]{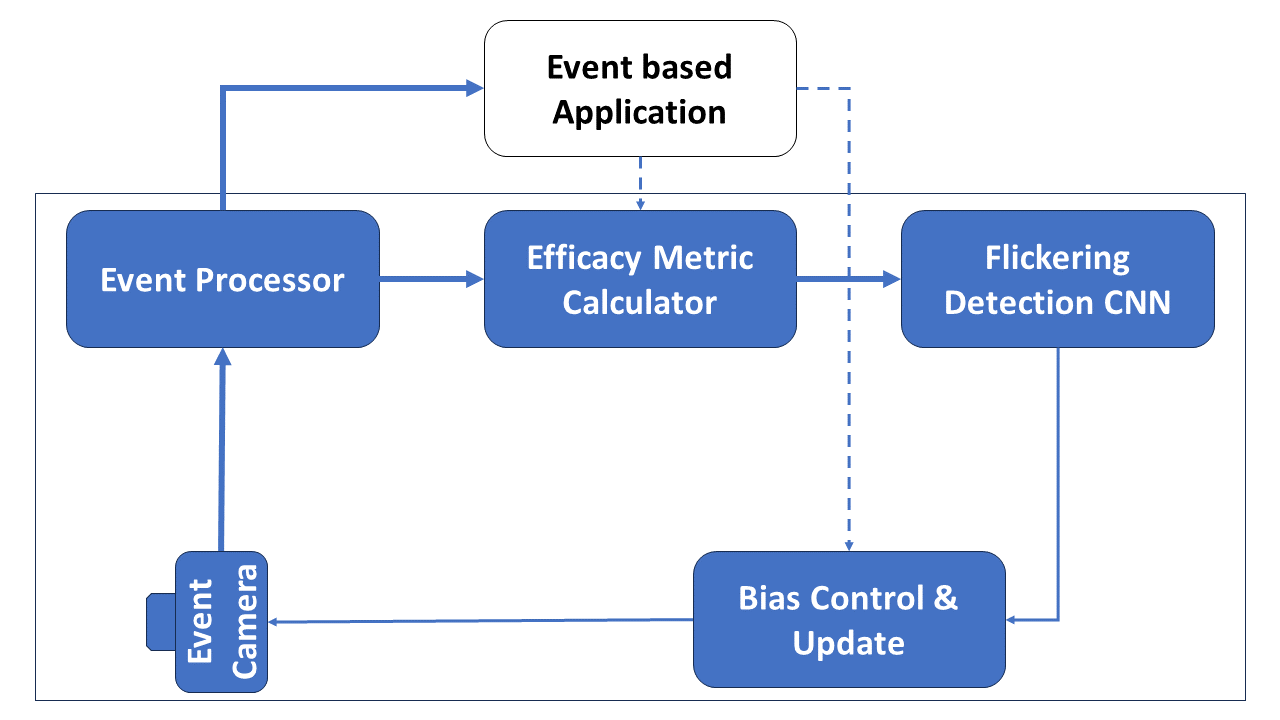}
	\caption{Auto-biasing block diagram}
	\label{fig_1}
\end{figure}

The block diagram shown in Figure \ref{fig_1} illustrates the auto-biasing process. As can be observed, at the beginning, the raw stream of events needs to be processed. In this paper, this processing is achieved by converting them into frames since, firstly, the flickering can be detected more clearly in frames than in the stream of raw events. Secondly, there is a wide range of mathematical metrics designed to assess image quality within traditional image processing methods. In this context, it's important to note that the average gradient, which is used to measure the effectiveness of auto-biasing, as discussed in \cite{m1}, is specifically designed for frames and may not apply to other data types. Thirdly,in the realm of computer vision, a majority of the current applications such as object detection networks, utilize networks that have been trained on images and frames. 

After the raw events have been processed, those are then ready for use in a specific application and subjected to evaluation using mathematical metrics in the subsequent phase. The effectiveness of this structure is showcased by using a modified YOLO V3 for face detection \cite{m23} as an example application. To quantify the system's performance, we leverage YOLO's confidences in detecting objects and faces, along with average gradient as a mathematical metric and another metric measuring the method's success in the time domain. In order to detect whether or not the frame contains flickers, it is next fed into a binary CNN classifier. 

It should be emphasized that AG is the primary metric in this study. Face detection confidences are calculated before and after bias modification to confirm the effectiveness of the proposed structure in real-world applications. Importantly, the algorithm is not limited to face detection and is designed to be applicable to any event-based application. The goal is to test the autobiasing's effectiveness with metrics specific to each application. Consequently, the face detector block and its outputs, which are shown as dashed lines in Figure \ref{fig_1}, are not part of the main structure and could be substituted with other applications.

The bias control and update unit receives the outcome of the classification along with the mathematical metric and feedback from the application. Then this block determines the changes required to occur in bias amounts and updates the camera settings based on its decision. Following sections explain the aforementioned blocks in detail.

\subsection{Events Processor}
The first step in automatically reducing flickering is detecting it, and to do this, a binary classifier is required to continuously track the output of the event camera to identify flickers as soon as they are present in the raw data. As was already mentioned, the proposed algorithm is based on frames rather than stream of events for three reasons: first, given the availability of many CNN classification algorithms developed for frame analysis, detecting flickering in frames becomes a simple task. Second, the efficacy of the proposed approach for flickering reduction must be measured. Since event cameras are a new technology, there are few metrics for evaluating their outputs \cite{m1}. Converting these outputs into frames allows for the use of established assessment algorithms. However, the traditional image processing metrics need careful adaptation to suit event frames. For tasks like flickering mitigation, this approach effectively tackles the complex assessment of event data. Third, although there is a growing trend among researchers to explore direct event processing using methods such as Spiking Neural Networks (SNNs) and Graph Neural Networks (GNNs), many current computer vision applications still rely on CNNs or other AI backbones trained with frames. Therefore, ensuring compatibility with these networks is a reliable way to fully utilize their benefits.

In this project, the term 'frame' refers to two dimensional matrices that match the camera's resolution, with each matrix's columns and rows corresponding to the number of pixels in the camera's width and height. The elements of these matrices are composed of aggregated pixel polarities collected over specified time intervals. The algorithms that converts the output of the cameras into frames, receives the stream of events and produces frames periodically. The algorithm is designed in a way that both the number of frames generated per second and the accumulation time of the events included in the frame can be specified \cite{m3}.

There are two important points to consider. Firstly, flickering becomes a concern if it persists over a considerable period of time. Secondly, in event cameras, the generated frames are not snapshots of fixed intervals; instead, they capture all events occurring between two consecutive frames, making it easier to record flickering effects. Therefore, generating and sending 10 frames per second to the flickering detector is sufficient and does not impose a significant computational load on the processor.

To address flickering that may exist in one frame but not the next, each frame is multiplied by a decay factor of 0.1 and then added to the next frame generated from pure events. This approach enhances the detection of flickering across multiple frames.

\subsection{Event-based Application}
It is important to remember that the purpose of autobiasing and flicker elimination is to enhance the functionality of an event based application. This application could be a smart home sensing system, an in-cabin or out-of-cabin vehicle monitoring system, or any other vision system. Therefore, the intended application should also demonstrate the autobiasing's efficacy; otherwise, there would be no benefit in employing it.

This research employs a YOLO V3-based face detection algorithm, as documented in \cite{m23}, to explain the benefits of autobiasing for flicker mitigation within the scope of event-based applications. Utilizing a novel architecture gated recurrent YOLO (GR-YOLO) for multi-object detection and tracking, particularly focusing on faces and eyes, the GR-YOLO incorporates a fully convolutional gated recurrent unit (GRU) layer to address the temporal sparsity of events. The performance of the face detection network was assessed quantitatively on a manually annotated dataset from a Prophesee event camera and tested qualitatively on real event driving data. To overcome the limited availability of event-based data, the method \cite{m23} used synthetic event camera dataset, Neuromorphic-Helen (N-Helen), generated from existing RGB datasets with facial landmark annotations. This dataset is designed to facilitate the training of event-based DMS models. The network architecture of \cite{m23} produces comprehensive bounding box predictions for objects, with a specific focus on faces and eyes. For each prediction, the model generates essential parameters: the coordinates of the bounding box center (\emph{bx} and \emph{by}), the width and height of the bounding box (\emph{bw} and \emph{bh}), the objectness probability (to), and class probabilities (\emph{p1} and \emph{p2}) indicating the likelihood of belonging to each of the two classes (face and eye). These predictions are calculated using 1x1 convolutions prior to the YOLO detection layers, which operate at two scales (8x8 and 16x16), resulting in a total of 960 predictions (3 anchor boxes at each of the 8x8 and 16x16 cells). Following predictions, the model employs post-processing techniques, including filtering based on objectness scores and non-maximum suppression, to refine and select the most accurate and relevant bounding box predictions. The overall objective of this architecture is to achieve efficient and accurate object detection, emphasizing faces and eyes in particular. For more information on the network please refer \cite{m23}.

When flickering occurs, successful detection may be hindered by the loss of facial details and features. The system will automatically adjust its bias settings until the face detection network accurately detects the face again. Furthermore, confidence values in the network's output will be assessed before and after bias tuning to illustrate system functionality through improvements in confidence levels.

\subsection{Efficacy Metrics Calculation Block}
This block is intended to assess the quality of the frames once they are generated. The first benefit of this assessment, which is employed in this research, is to compare the quality of the frames before and after each loop of bias adjustment in order to evaluate the effectiveness of the auto-biasing. The second benefit of this block is that it can generate controlling signals in the feedback loop to modify the biases based on quality changes after each loop is complete.

\subsubsection{Average Gradient (AG)}
The nature of the flickers and how they appear in the frames must be taken into account in order to develop a suitable metric for measuring the amount of flickering in frames. Given that the flickers cause numerous sharp variations in the intensity of the pixels, edge detection would be an ideal technique to track these variations. Canny edge detector \cite{m5}, Sobel operator \cite{m34}, Prewitt operator \cite{m35}, Laplacian of Gaussian (LoG) \cite{m36}, Robert's cross operator \cite{m37}, and Frei-Chen operator \cite{m38} are the most frequently used methods for edge detection. Among these, Canny edge detection is a common technique that offers benefits in edge tracking, precise edge localization, and automatic thresholding \cite{m5}. 

The Canny edge detection approach emphasizes the importance of gradient magnitude in edge recognition by using it to identify edges in an image \cite{m5}. Other than that, the method proposed in \cite{m6}, explores steerable filters, which can calculate responses in any direction and are primarily used for edge and texture analysis. It underscores the crucial role that gradient information plays in image analysis. Deriche additionally proposes an alternative to the edge detection method employed in \cite{m5} and highlights the significance of gradient magnitude calculation in the edge detection procedure \cite{m7}. Sobel introduces an operator in \cite{m8} that is now referred to as the Sobel operator and is a key tool for determining gradient magnitude in image processing.

These methodologies formed the basis of our adoption of the average gradient magnitude (AG) computation to assess the impact of bias adjustments on event-based frames in a different project \cite{m1}. Gradients are used to indicate the changes in picture intensity in image processing applications. Estimating the gradient at each individual pixel and then repeating the procedure for the entire image is one method for locating edges in an image. To calculate the AG, the gradient along the x and y axes must first be measured, and the average gradient value can then be calculated across all pixels, as the formula \ref{eq_1} indicates:

\begin{equation}
  AG = \frac{1}{MN} \sum_{i=1}^{M} \sum_{j=1}^{N} \sqrt{\left(\frac{\partial f}{\partial x}\right)^2 + \left(\frac{\partial f}{\partial y}\right)^2}
  \label{eq_1}
\end{equation}

where $M$ and $N$ are the number of pixels in the x and y axes of the image $f$, respectively. The $AG$ will increase as the number of edges on an image increases. So, the presence of flickering in an image will increase its $AG$. The effectiveness metric calculator in Figure \ref{fig_1} calculates the aforementioned formula for each generated frame. After the auto-biasing mechanism is activated to reduce flickering, it is expected that the amount of the calculated $AG$ would decrease. 

As the first auto-biasing approach for flickering mitigation in event cameras, in this paper an universal metric is adopted, AG, that applies to all frames. This metric is a crucial component of the autobiasing system for flicker removal, evaluating the degree of flickering within each frame. Application-specific metrics can easily supplement the Average Gradient (AG) for particular applications such as detecting objects, driver monitoring, blink counting, etc. without necessitating changes to the proposed autobiasing strategy's main structure or other components. In this research, which employs a face detection network to demonstrate effectiveness, additional metrics specific to this application are also utilized and will be explained in the following part.

\subsubsection{Application Specific Assessment Metrics}

\textbf{YOLO Confidences:}
The specific YOLO metrics evaluated in this particular case are face detection confidence and confidence in any object detection. To demonstrate the impact of autobiasing, these factors are measured both before and after the process. As mentioned in the "Event-based Application" section, the any object confidence reflects the likelihood of an object being present in the frame, while the face confidence indicates the probability of a face being present in the same frame. Both of these confidences are expected to increase after autobiasing.

\textbf{Face Detection Success in the Time Domain:}
Another metric is also adopted from former studies \cite{m32}. The main idea of this method is comparing the number of frames with a detected face in a second with the total number of frames captured in that second. A proper autobiasing, must lead to an increment in this metric.
The face detection success metric is calculated as follows:

\begin{equation}
  Face Detection Success = \frac{F_{detected}}{F_{total}}
\end{equation}

The number of frames containing a detected face is denoted by $F_{detected}$, whilst the total number of frames in a second is indicated by $F_{total}$.

\subsection{Flickering Detection Classifier}
\subsubsection{Network Architecture}
\begin{figure}[h]
	\centering
		\includegraphics[width=.85\linewidth]{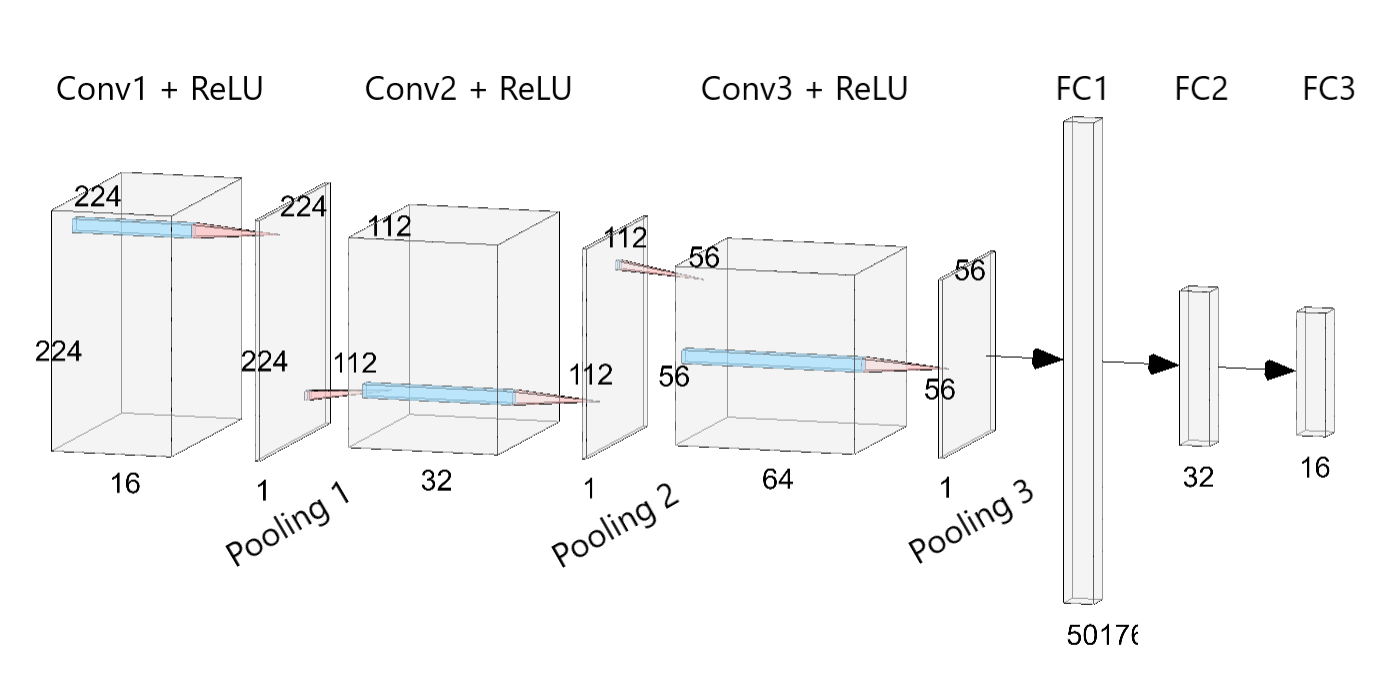}
	\caption{Architecture of the flickering classification CNN}
	\label{fig_3}
\end{figure}

As a result of the advantages of convolutional neural networks in image analysis applications, a CNN is used to split the generated frames into two categories: flickering frames and frames without flickering. The network, as shown in Figure \ref{fig_3} is made up of three convolutional layers and a fully connected layer, as is being discussed subsequently:

First convolutional layer starts with a 2D convolutional layer that accepts the input frames generated in the event processor, outputs 16 channels, uses a 3x3 kernel, has a stride of 1, and padding of 1. The padding of 1 ensures that the spatial dimensions of the input remain unchanged after this convolution operation. The activation function which introduces non-linearity to the model is a rectified linear unit (ReLU). A 2x2 max pooling that halves the spatial dimensions of the input both in height and width is added to the layer. Second convolutional layer is similar to the first layer, but this time it accepts an input with 16 channels from the previous layer and outputs 32 channels. Third convolutional layer accepts input with 32 channels and outputs 64 channels. Fully connected layers start with a liner layer that accepts a flattened input of size 64x28x28 and outputs 32 features. The dimension 64x28x28 assumes the input frame to the network is of size 224x224 and after passing through three 2x2 max pooling layers it becomes 28x28. Followed by two more linear layers that reduce the dimensionality to 16 and finally to 2, since it is designed for a binary classification task.

The training of the neural network was conducted over 30 epochs using the Adam optimizer. The optimizer was configured with a learning rate of 0.001, beta coefficients of 0.9 and 0.999, an epsilon value of 1e-08, and no weight decay. Cross-Entropy Loss was employed as the loss function to effectively manage the classification task.

\subsubsection{Dataset}
All datasets used in this study were balanced and captured under natural office lighting at various times of day, including both well-lit and low-light conditions, using a Metavision EVK4 HD event camera \cite{m9}. The recording setup involved positioning the event camera approximately 1.5 meters away from the subjects. The subjects included human participants seated in front of the camera, as well as static objects such as office equipment, and dynamic objects like revolving motors. The dataset is comprised of 3560 frames for training and 552 frames each for evaluation and testing. Two categories were defined: frames labeled as flickering and frames labeled as non-flickering. For the non-flickering category, event data was captured in a flicker-free environment, ensuring stable lighting conditions. The flickering category was created by introducing flicker using infrared (IR) emission sources at varying frequencies. This method allowed us to generate controlled flickering conditions, ensuring that the ground truth for training the network was accurate. Ground truth labels were assigned based on the controlled conditions: non-flickering labels were assigned to frames captured without the IR source, while flickering labels were assigned to frames with IR-induced flickering. Figure \ref{fig_4} provides visual samples from each category.

\subsubsection{Training Procedure}
\begin{figure}[h]
	\centering
		\includegraphics[width=.80\linewidth]{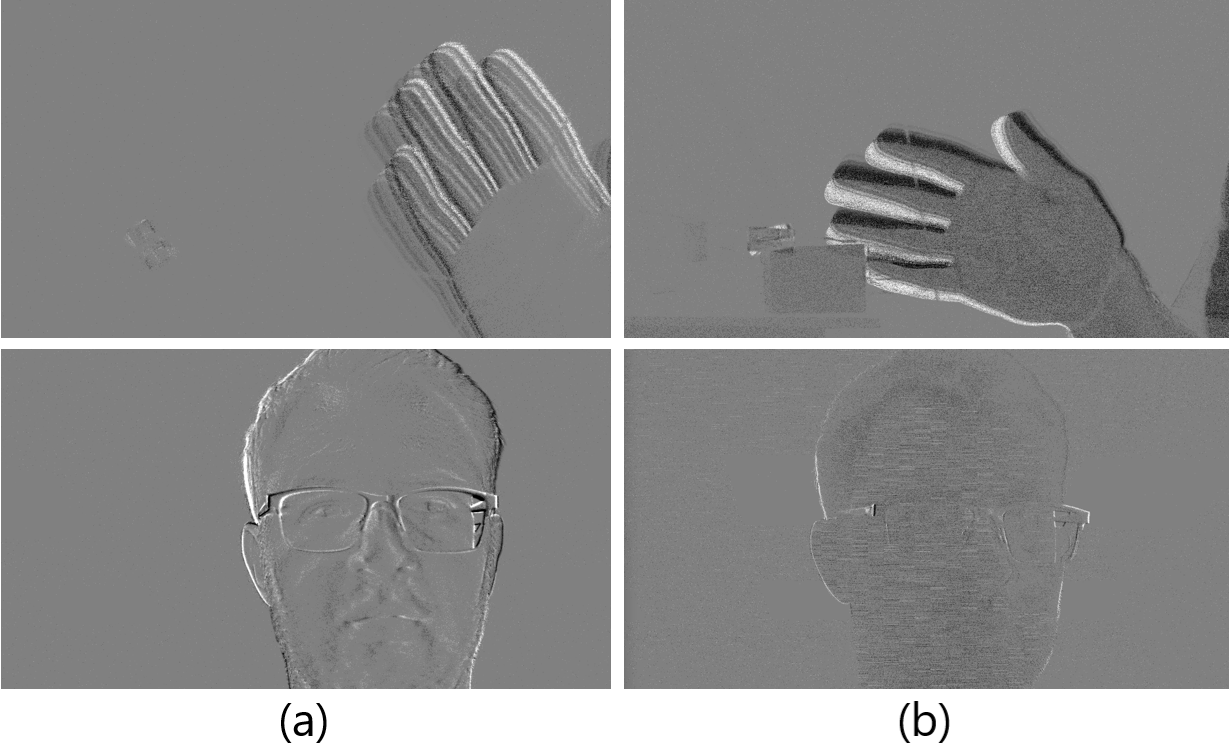}
	\caption{Samples of frames: a) without flickering, b) with flickering}
	\label{fig_4}
\end{figure}

The two categories had significant differences, as seen in Figure \ref{fig_4}, and the network performed very well during the test phase. The achieved test accuracy after 10 training iterations is 99.86\%.

\subsection{Bias Control and Update Unit}
This block processes three input parameters, to produce two output parameters. Two of the inputs are provided by the other blocks and include the AG value, and the flickering status of each generated frame. The third input is the bias\_fo value read from the camera. This block's primary objective is to determine a new value for the bias, which is the first output and is written to the camera. The efficacy of the auto-biasing is demonstrated by the second output. 

Despite the very high accuracy of the classification network, it is still important to design the process so that inaccurate predictions have the least negative effects on the system's overall functionality. To do this, the control unit decides about updating the bias value every second after receiving information on all frames generated in the last second, as opposed to modifying the bias frame by frame. Every second, this unit explores the number of frames from each class and determines whether or not there is flickering depending on the results of the comparison. If the system detects flickering, it modifies the bias fo value to decrease the pixels' low pass filter bandwidth and minimize the flickering effect. The event camera runs slowly at lower bandwidths, thus it is required to raise the bandwidth if the system cannot verify that there is flickering. The bias fo amount in the camera used for this project needs to be reduced in order to reduce the bandwidth, and vice versa. It is important to consider that the bias range is constrained to values between (-35 and 55), and that the bias values given to the camera can't exceed outside of this range. In this project the bias values are modified in steps of size 5. 

As long as flickering is detected, the system keeps reducing the bias value by that step size. If flickering is still present after the bias has been set to its minimum value, it means that bias tuning will not be able to eliminate the flickering, and another strategy has to be established. The process is slightly different to raise the bias value, and the bias rises if the system cannot detect flickering for 10 seconds in row. This procedure is carried out to avoid the system from getting trapped in a loop where it finds flickering, lowers the bias, removes flickering, mistakenly believes the source of the flickering has been removed, raises the bias value, and then detects flickering again.

\begin{figure}[h]
	\centering
		\includegraphics[width=.85\linewidth]{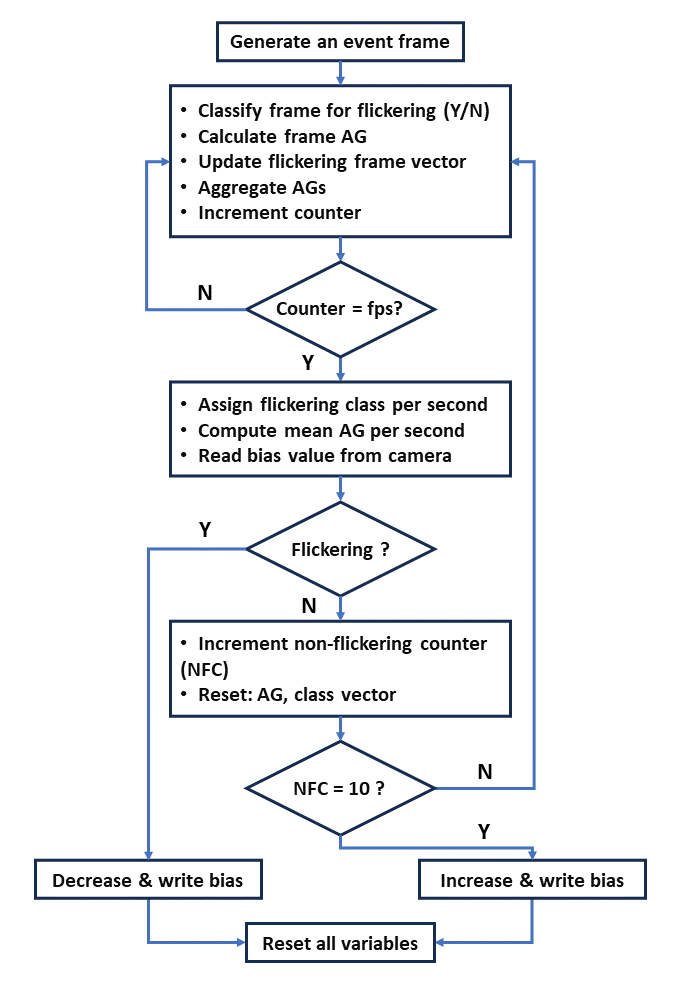}
	\caption{Bias control and optimization process: example of face detection}
	\label{fig_5}
\end{figure}

Every second, once the system has assessed if there is flickering, the mean value of the AGs for all of the frames in the last second is calculated, along with the new bias value. This value should continue to decrease with each bias update in the presence of flickering, proving the efficacy of the auto-biasing. Figure \ref{fig_5}, illustrates the bias control and optimization process.

It's important to note that selecting different time intervals does not impact the core process of autobiasing. In this study, a 1-second interval is selected to detect flickering presence and a 10-second interval to check the flickering source persistence after autobiasing.

\section{Results and Discussion}
To demonstrate the effectiveness of the proposed system and its advantages over other flicker mitigation algorithms, we conducted tests simulating real-world conditions across various lighting scenarios, including diverse flickering frequencies in both illuminated and dark environments. All results presented are derived from real-time autobiasing. The proposed algorithm operates directly with live data from the camera, eliminating the need for saving or preprocessing data in memory. Consequently, the system is neither trained nor tested on static datasets; instead, flickering mitigation occurs directly on the data streaming from the event camera. All the tests were conducted under natural light in the office and there was no artificial light on in the test room. To measure the amount of the light a SEKONIC C-500 light meter was used. 5 different people (whose informed consent was obtained) sat in the distance of 50 cm away from the Prophesee IMX 636 event camera and a flickering was generated using 4 IR LEDs. To drive the flickering LEDs and changing their frequency an AC signal generator was used. In well-lit conditions, the tests were carried out in an office environment with around 1000 lux, while in low-light conditions, the illumination was less than 20 lux. 

The experiments for this study were carried out on a high-performance computing system powered by an Intel(R) Core(TM) i7-10870H CPU, which operates at a base frequency of 2.20 GHz, capable of reaching 2.21 GHz. The system includes 63.8 GB of usable RAM and an NVIDIA RTX 3080 GPU, enhancing our computational capabilities for deep learning tasks. The development and test environment featured Python and PyTorch. Additionally, Prophesee libraries were employed to interface with event-based camera systems, allowing for effective processing of incoming data. Running the proposed algorithm on low-power devices is beyond the scope of this research; however, it would require careful quantization and the selection of a device capable of processing data from the event camera in real time.

The effectiveness of autobiasing is demonstrated through its impact on enhancing camera performance, as evidenced in this project by improvements in face detection confidence scores and AG values. It is crucial to focus on the amount of changes in these metrics before and after applying autobiasing. This approach is important because the absolute values of these metrics can vary widely among individuals and under different testing conditions. Thus, assessing the improvements offers a more consistent measure of autobiasing's efficacy. 

\begin{figure}[htp]
\centering
\includegraphics[width=.3\textwidth]{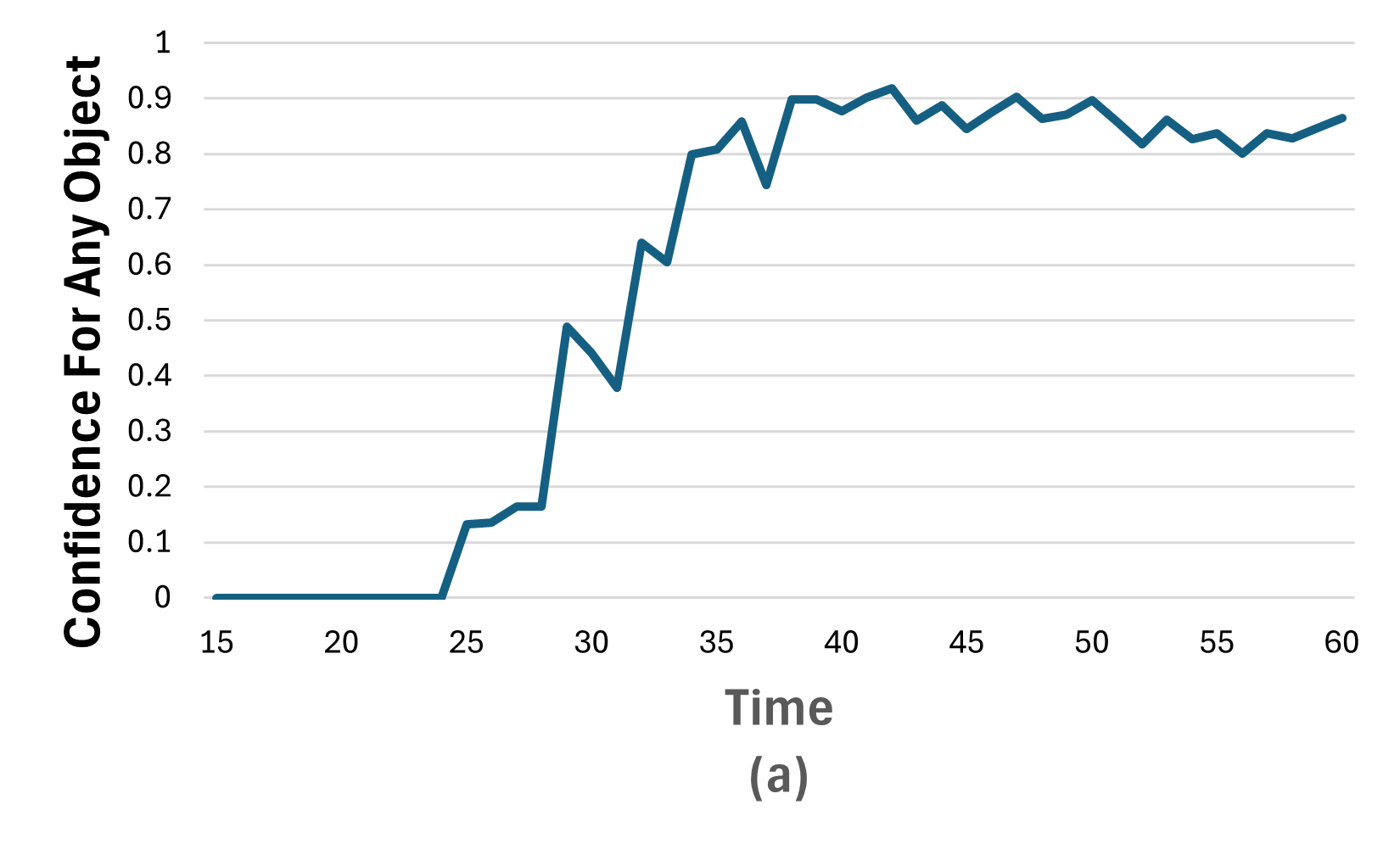}\hfill
\includegraphics[width=.3\textwidth]{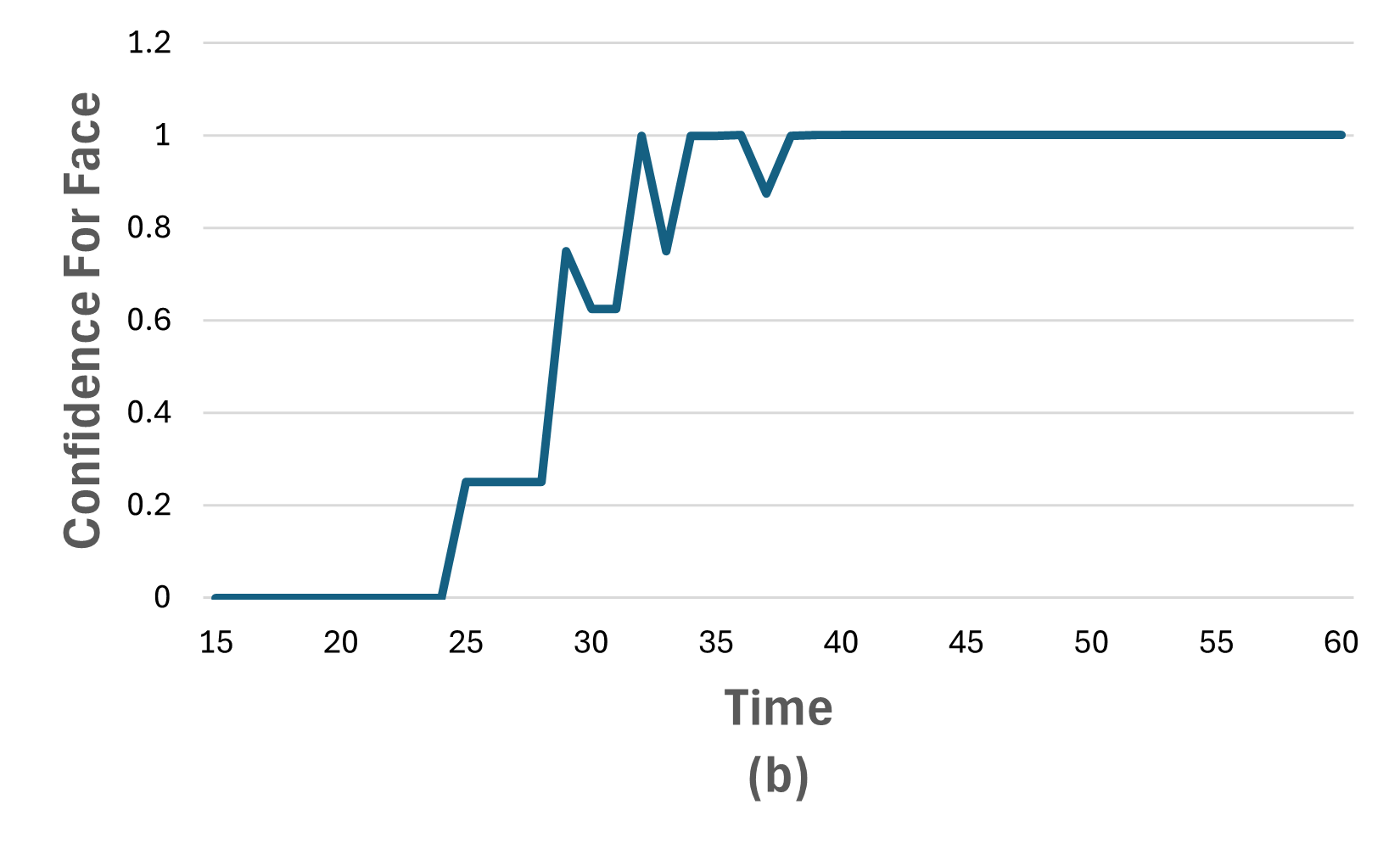}\hfill
\includegraphics[width=.3\textwidth]{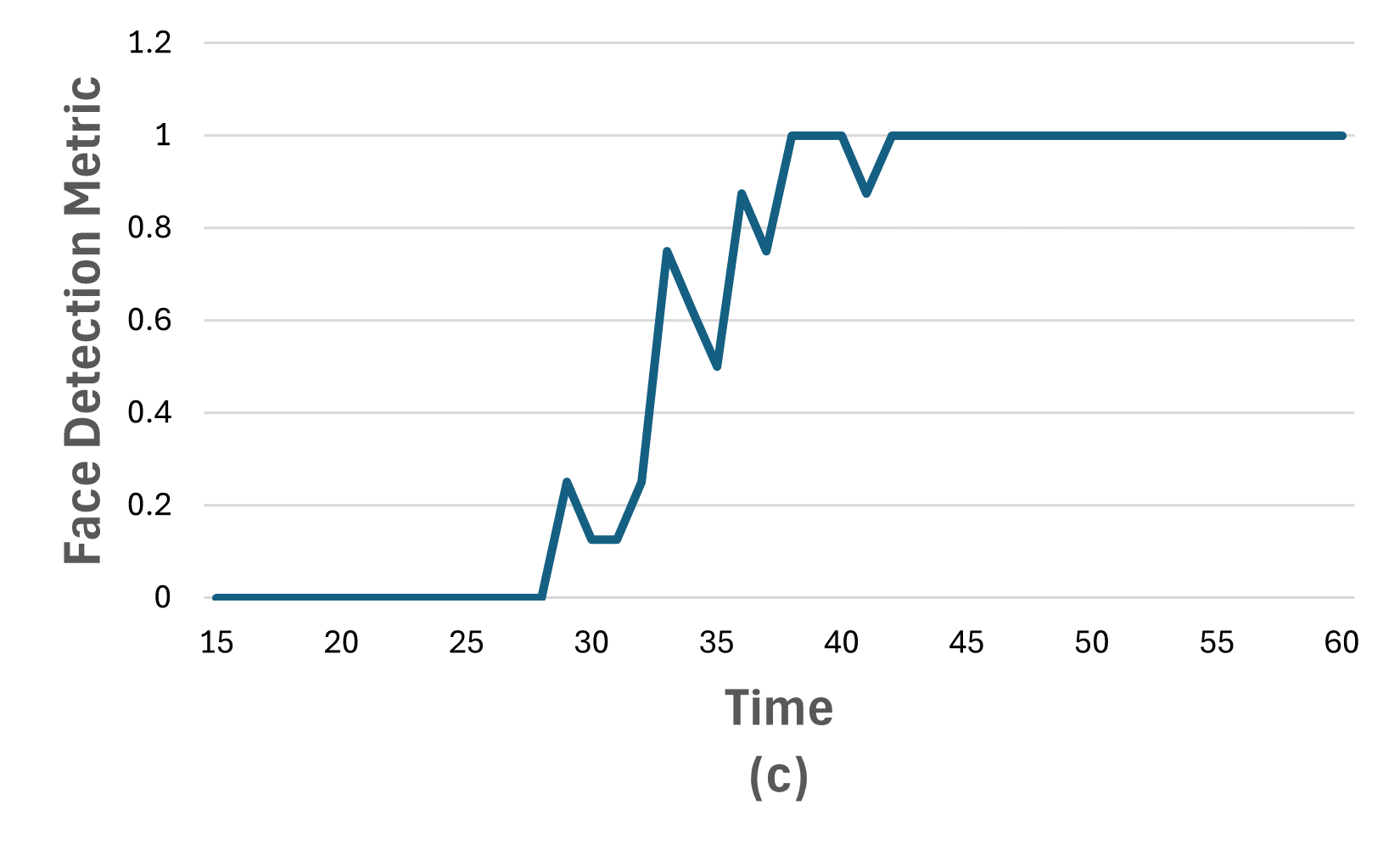}\hfill
\includegraphics[width=.3\textwidth]{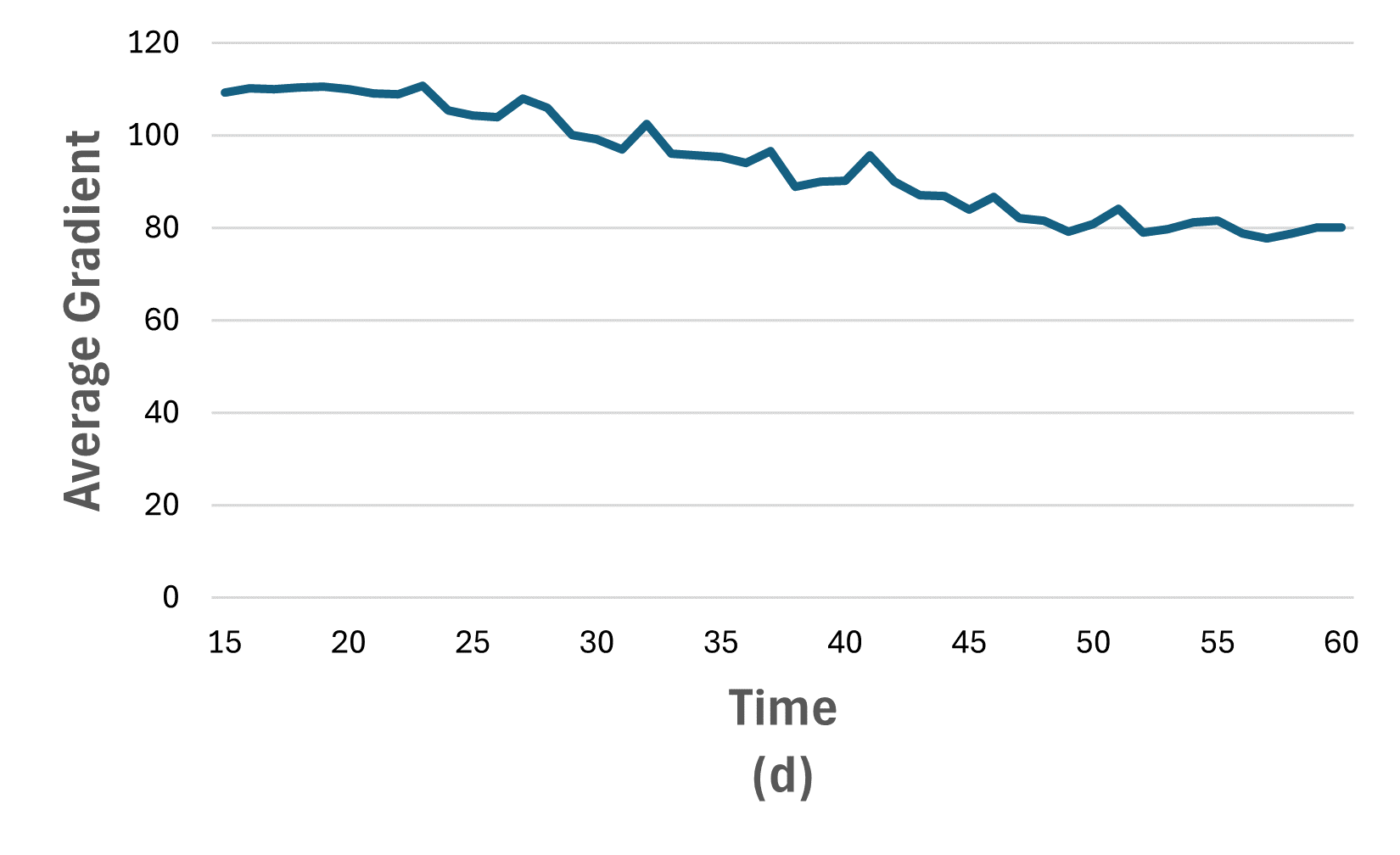}

\caption{Positive effects of autobiasing on metrics over time under high light with 25 Hz flickering: a) Rise in confidence for object presence. b) Rise in confidence for face detection. c) Higher ratio of frames with detected faces compared to all frames. d) Reduction in average gradient, signifying enhancement.}
\label{fig_7}
\end{figure}

The average changes in four metrics, observed 60 seconds after the autobiasing beginning, are presented in Tables I and II. The metrics are Average Gradient (AG), YOLO confidence in any object detection, YOLO confidence in face detection, and the percentage of the frames with detected face compared to all of the frames in 1 second. Figure \ref{fig_7} illustrates the modification of these metrics during the autobiasing process for a sample subject in the high-light condition with a flickering frequency of 25 Hz. The graphs show improvements across all metrics over time. Improvement is characterized by a decrease in average gradient and an increase in other metrics.

\begin{figure}[h]
	\centering
		\includegraphics[width=.8\linewidth]{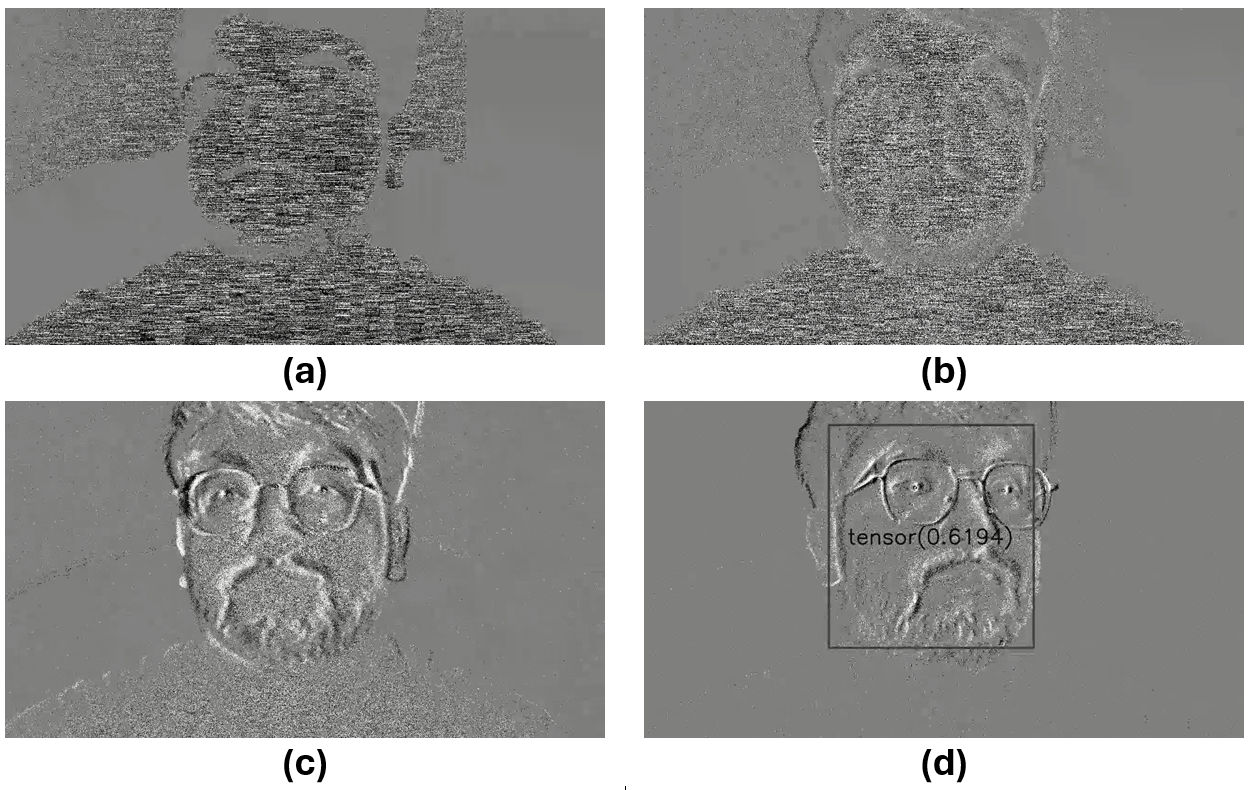}
	\caption{Autobiasing progress frames: a) Before autobiasing, face undetectable. b) Autobiasing initiation mitigates some flickering. c) Flickering reduced, face visible but not detected by YOLO. d) Autobiasing completion, flickering eliminated, face detected by YOLO.}
	\label{fig_8}
\end{figure}

In addition to the quantitative analysis provided in tables I and II and Figure \ref{fig_7}, a qualitative assessment is depicted in Figure \ref{fig_8} to demonstrate the efficacy of the proposed method. Four frames illustrating flickering mitigation in low-light conditions are presented. Initially, the flicker obscures the image, but as autobiasing initiates in the second frame, flickering is gradually mitigated. Next frame shows further improvement, with the subject's face becoming progressively clearer and more distinguishable, peaking in the final frame where the face is easily visible and detected by YOLO with a confidence of 62\%.

\begin{table}[t]
    \centering
    \caption{Changes in key metrics following autobiasing under varying lighting (high-light: Lux $>$ 1000; low-light: Lux $<$ 20) and various flicker frequencies.}
    \adjustbox{width=\columnwidth}{
        \begin{tabular}{|c|c|c|c|c|}
            \hline
            \multirow{2}{*}{Frequency (Hz)} & \multicolumn{2}{|c|}{Confidence (\%)} & \multirow{2}{*}{Frames with detected face(\%)} & \multirow{2}{*}{AG (\%)} \\
            \cline{2-3}
            & Any Obj. & Face & & \\
            \hline
            \multicolumn{5}{|c|}{Lux $>$ 1000} \\ 
            \hline
            25 & +74 & +89 & +88 & -42 \\
            \hline
            50 & +73 & +85 & +82 & -41 \\
            \hline
            150 & +70 & +85 & +83 & -44 \\
            \hline
            300 & +72 & +82 & +75 & -39 \\
            \hline
            500 & +57 & +54 & +73 & -25 \\
            \hline
            \multicolumn{5}{|c|}{Lux $<$ 20} \\ 
            \hline
            25 & +42 & +52 & +51 & -54 \\
            \hline
            50 & +29 & +40 & +36 & -51 \\
            \hline
            150 & +38 & +52 & +50 & -42 \\
            \hline
            300 & +44 & +61 & +58 & -64 \\
            \hline
            500 & +38 & +61 & +60 & -57 \\
            \hline
        \end{tabular}
    }
    \label{tbl_1}
\end{table}

\subsection{Comparative Analysis of Event Flicker Mitigation Techniques}
As outlined in the literature review section, each proposed method has its own advantages and shortcomings. However, from our perspective, the primary challenges in the flickering mitigation from event data include: the flickering frequency, the ability of system to work in real-time mode, the requirement for extra hardware or software resources, and finally a mechanism to monitor the effect of the flickering mitigation in the main event-based application. Table III summarizes the performance of each method, considering these points.

\newcolumntype{C}[1]{>{\centering\arraybackslash}p{#1}}

\begin{table*}
\centering
\caption{Comparison of Various Event Flickering Mitigation Methods}
\begin{tabular}{|C{2.75cm}|C{2.5cm}|C{2.25cm}|C{8cm}|}
\hline
\textbf{Method} & \textbf{Test Freq. (Hz)} & \textbf{Mode} & \textbf{Requirement for Additional Hardware/Software} \\
\hline
Comb filter \cite{m21} & 50 & Non-Real-Time & Requires additional filters \\
\hline
PINK \cite{m22} & 50 \& 60 & Real-Time & Requires multiple filters and has a complex structure \\
\hline
Proposed Method & 25 to 500 & Real-Time & No additional resources; utilizes built-in biases \\
\hline
\end{tabular}
\end{table*}

Considering flickering frequency, the proposed method has been tested across a broader range of frequencies. It operates in real-time and does not require additional hardware or software resources to implement a filter. Moreover, as previously stated, methods that remove flickering could also inadvertently remove real data. Thus, flickering mitigation must be dynamic and adaptive to avoid negatively impacting the primary event-based application. Uniquely, our method not only monitors face detection as a representative application but is also capable of receiving feedback from other applications.

\section{Conclusion and Future Work}
Event cameras are capable of being utilized for many different purposes. They operate effectively across various lighting conditions but can encounter challenges when these conditions change. One of the significant challenges affecting the performance of event cameras stems from flickering light sources. This issue is more pronounced in event cameras due to the similarity in amplitude between flickering signals and the main signal. This flickering imposes significant computational and storage burdens on output data. This research addresses the issue of flickering in event camera outputs by introducing a novel system that automatically detects and mitigates flickering. It utilizes the adjustable bias settings of event cameras, incorporating an automatic flicker detection and removal system. Flickering is detected by a CNN trained on local data, and upon detection, the bias\_fo is dynamically adjusted to reduce the flickering until the CNN can no longer detect it. The average gradient is utilized as a mathematical metric to demonstrate the quality improvement in the event camera output after autobiasing. Additionally, YOLO-based face detection is applied to showcase the potential benefits of autobiasing, aiming to enhance the algorithm's ability to detect faces in flickering conditions. The proposed system was tested with five distinct flickering frequencies, ranging from 25 Hz to 500 Hz, in both well-lit (over 1000 lux) and low-light (under 20 lux) conditions. In well-lit condition, notable improvements were observed across the explored metrics: the average gradient improved by 38.2\%, YOLO confidence for any object by 69\%, confidence for face by 79\%, and the percentage of frames with detected faces at each second by 80.2\%. Similarly, great enhancements were observed in low-light conditions, with improvements of 53.6\% in average gradient, 38.2\% in YOLO confidence for any object, 53.2\% in confidence for face, and 51\% in the percentage of frames with detected faces at each second.

The proposed method can be adopted by various applications and effectively mitigates the flickering impact by modifying a single bias, bias\_fo. Later research will explore the possibility of tuning all biases for event based computer vision applications.


\end{document}